\documentclass[technote,compsoc]{IEEEtran}

\usepackage{graphicx}
\usepackage{subfig}
\usepackage{caption}
\usepackage{amssymb}
\usepackage{pifont}
\usepackage{amsfonts}
\usepackage{amsmath}
\usepackage{booktabs}
\usepackage{tabularx}
\usepackage{multirow}
\usepackage{xcolor,colortbl}
\usepackage{color}
\usepackage[ruled, vlined]{algorithm2e}

\ifCLASSOPTIONcompsoc
  \usepackage[nocompress]{cite}
\else
  \usepackage{cite}
\fi

\definecolor{Gray}{gray}{0.85}
\definecolor{mycolor}{HTML}{00D1B2}

\usepackage{array}
\newcolumntype{x}[1]{>{\centering\arraybackslash\hspace{0pt}}p{#1}}

\newcolumntype{a}{>{\columncolor{Gray}}c}

\usepackage[pagebackref=true,breaklinks=true,letterpaper=true,colorlinks,bookmarks=false]{hyperref}

\begin{document}

\title{Reducing Spatial Labeling Redundancy for Semi-supervised Crowd Counting}

\author{\normalsize{Yongtuo Liu,
	    Sucheng Ren,
	    Liangyu Chai,
	    Hanjie Wu,
	    Jing Qin,
	    Dan Xu,
        and Shengfeng He,~\IEEEmembership{Senior Member,~IEEE}}
\IEEEcompsocitemizethanks{
\IEEEcompsocthanksitem Yongtuo Liu, Sucheng Ren, Liangyu Chai, Hanjie Wu, and Shengfeng He are with the School of Computer Science and Engineering, South China University of Technology, Guangzhou, China. E-mail:  csmanlyt@mail.scut.edu.cn, oliverrensu@gmail.com, 
icepoint1018@gmail.com, 
cshanjiewu@gmail.com, hesfe@scut.edu.cn.
\IEEEcompsocthanksitem Jing Qin is with the Department of Nursing, Hong Kong Polytechnic University. E-mail: harry.qin@polyu.edu.hk.
\IEEEcompsocthanksitem Dan Xu  is with the Department of Computer Science and Engineering, Hong Kong University of Science and Technology. E-mail: danxu@cse.ust.hk.
}
}

\IEEEtitleabstractindextext{%
\begin{abstract}
Labeling is onerous for crowd counting as it should annotate each individual in crowd images.
Recently, several methods have been proposed for semi-supervised crowd counting to reduce the labeling efforts.
Given a limited labeling budget, they typically select a few crowd images and densely label all individuals in each of them.
Despite the promising results, we argue the None-or-All labeling strategy is suboptimal as the densely labeled individuals in each crowd image usually appear similar while the massive unlabeled crowd images may contain entirely diverse individuals.
To this end, we propose to break the labeling chain of previous methods and make the first attempt to reduce spatial labeling redundancy for semi-supervised crowd counting.
First, instead of annotating all the regions in each crowd image, we propose to annotate the representative ones only.
We analyze the region representativeness from both vertical and horizontal directions, and formulate them as cluster centers of Gaussian Mixture Models.
Additionally, to leverage the rich unlabeled regions, we exploit the similarities among individuals in each crowd image to directly supervise the unlabeled regions via feature propagation instead of the error-prone label propagation employed in the previous methods.
In this way, we can transfer the original spatial labeling redundancy caused by individual similarities to effective supervision signals on the unlabeled regions.
Extensive experiments on the widely-used benchmarks demonstrate that our method can outperform previous best approaches by a large margin.
\end{abstract}

}

\maketitle

\IEEEdisplaynontitleabstractindextext
\IEEEpeerreviewmaketitle

\section{Introduction}
Crowd counting has draw increasing attention in the community due to its essential role in social management, such as crowd monitoring, crowd congestion warning, and etc~\cite{li2014crowded, loy2013crowd, gao2020cnn}.
Benefiting from the powerful CNN architectures, lots of works have been proposed and advanced the performance of crowd counting.
Most of them are mainly dedicated to solving various challenges of crowd counting in a fully-supervised manner~\cite{zhang2016single, liu2019context, bai2020adaptive, wan2020modeling, lian2019density, shi2019revisiting}.
However, labeling for crowd counting is quite burdensome as we have to annotate each individual in crowd images.

To reduce the labeling efforts, we study crowd counting in a semi-supervised setting where only a small labeling budget is available.
Methods in this line can be mainly grouped into three categories:
(i) \cite{liu2018leveraging, sam2019almost, liu2020semi} leverage self-supervised constraints to learn a generic feature extractor from unlabeled crowd images.
(ii) \cite{sindagi2020learning, zhao2020active} introduce knowledge transfer to bridge the labeled and unlabeled data.
(iii) \cite{tan2011semi, loy2013semi, zhou2018crowd} exploit temporal labeling redundancy in crowd video scenarios.

\begin{figure}[t]
 \centering
 \setlength{\tabcolsep}{1pt}
 \includegraphics[width=.98\linewidth]{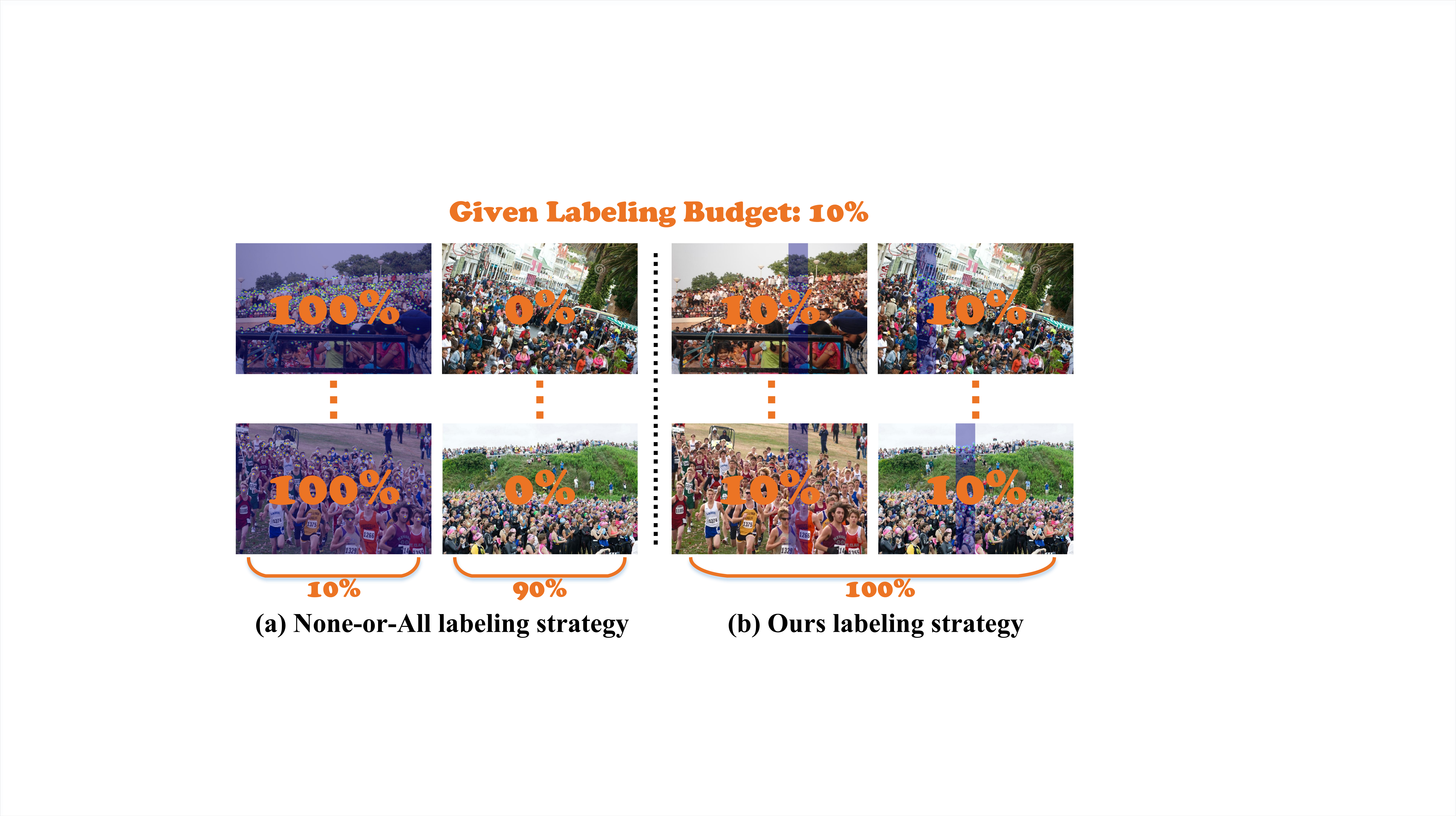}
 \caption{Given a limited labeling budget (e.g., 10\% of the entire dataset), all the previous methods adopt a None-or-All labeling strategy and select a few crowd images to densely label all the individuals which typically appear similar and lack diversity. Differently, we propose to break the labeling chain of previous methods and annotate the representative regions only in each crowd image.}
 \label{teaser}
 \end{figure}

Notwithstanding the demonstrated success of the above methods, they all view each crowd image as a minimum labeling unit and densely label all individuals in a limited number of crowd images.
The None-or-All labeling strategy is suboptimal considering the labeling burden and labeling redundancy in each crowd image.
(i) Compare to other computer vision tasks, the labeling burden of crowd counting mainly resides in each crowd image where hundreds of individuals may need to be annotated.
The existing methods try to alleviate the labeling burden by decreasing the number of labeled crowd images, which seems palliative for the crowd counting problem. 
(ii) Individuals in each annotated crowd image usually appear similar with lots of labeling redundancy as they are captured in the same crowd scene. This makes the annotated individuals lack diversity and cannot adapt to various crowd scenes, e.g., different camera perspectives, weather and illumination conditions.

To this end, we propose to break the labeling chain of previous methods and make the first attempt to reduce spatial labeling redundancy for semi-supervised crowd counting.
First, instead of annotating all the regions in each crowd image, we propose to annotate the representative ones only (see Fig.~\ref{teaser}).
We analyze the region representativeness from both the vertical and horizontal directions, and design a Multi-level Density-aware Cluster (MDC) Strategy to formulate the representative regions as cluster centers of Gaussian Mixture Models based on their multi-level density vectors.
In this way, our method can effectively reduce the spatial labeling redundancy in each crowd image and label more crowd images with various crowd scenes given the same labeling budget.
Additionally, to leverage the rich unlabeled regions, we further exploit the similarities among individuals in each crowd image to directly supervise the unlabeled regions via feature propagation in a Crowd Affinity Propagation (CAP) module.
The CAP module propagates crowd features of the unlabeled regions to update those of the labeled regions based on the feature affinities in the forward propagation.
Then in the backward propagation, the unlabeled regions can be directly supervised by the labels of the labeled regions via feature backpropagation.
After training, we can optionally remove the CAP module without performance degradation which makes it computationally free at the inference stage.
By the CAP module, we can transfer the original spatial labeling redundancy caused by the individual similarities to effective supervision signals and directly supervise the unlabeled regions without generating the error-prone pseudo labels.
Extensive experiments on widely-used benchmarks demonstrate our method outperforms previous approaches by a large margin.
For example, our method outperforms the best AL-AC~\cite{zhao2020active} by 9.4\%/8.1\% and 8.6\%/22.5\% for MAE/RMSE in the ShanghaiTec PartA and PartB datasets, respectively.

The contributions are summarized as follows:
\begin{itemize}
\item
We propose to break the labeling chain of previous methods and reduce the spatial labeling redundancy by annotating representative regions only for effective semi-supervised crowd counting.
\item
We analyze the region representativeness from both vertical and horizontal directions and formulate representative regions as cluster centers of Gaussian Mixture Models.
Furthermore, to leverage the unlabeled regions, we propose to exploit the similarities among individuals to directly supervise the unlabeled regions via feature propagation without the error-prone pseudo label generation.
\item
Extensive experiments show that our method can achieve state-of-the-art performance and outperform previous best approaches by a large margin.
\end{itemize} 

\section{Related Work}

\subsection{Crowd Counting}
Early methods for crowd counting are based on hand-crafted features (e.g., SIFT, Fourier Analysis, and HOG).
They estimate crowd counts by either direct regression~\cite{chen2012feature, idrees2013multi, lempitsky2010learning} or human parts detection~\cite{lin2010shape, wang2011automatic, wu2007detection}.
Recently, a lot of CNN-based methods have been proposed and advanced the performance of crowd counting.
Most of them mainly solve various challenges of crowd counting in a \emph{fully-supervised} manner, including large scale variations~\cite{zhang2016single, cao2018scale, li2018csrnet, liu2019adcrowdnet, sam2017switching, liu2019context, jiang2019crowd, luo2020hybrid, liu2020crowd}, attentive feature extraction~\cite{sindagi2019ha, jiang2020attention, miao2020shallow, zhang2019relational, zhang2019attentional}, label noises~\cite{bai2020adaptive, wan2020modeling}, empirical gaussian kernel~\cite{wan2019adaptive, ma2019bayesian, liu2020weighing}, estimation uncertainty~\cite{ranjan2020uncertainty, oh2020crowd}, structural constraints~\cite{lian2019density, shi2019revisiting, yan2019perspective, chai2020crowdgan}, and etc.
These methods require a great number of labeled data in the training process which are rather burdensome for crowd counting.

\subsection{Semi-supervised Crowd Counting}

Recently, several methods are designed to learn a crowd counter with a limited labeling budget.
They can be mainly grouped into three categories as follows:

\emph{Self-supervised Constraints}~\cite{liu2018leveraging, sam2019almost, liu2020semi}:
Liu et al.~\cite{liu2018leveraging} propose to exploit unlabeled data by ranking cropped patches according to their containment relationships.
Sam et al.~\cite{sam2019almost} extract useful feature representations by learning a Grid Winner-Take-ALL (GWTA) autoencoder from unlabeled crowd images.
Liu et al.~\cite{liu2020semi} propose to leverage surrogate tasks with IRAST constraints to train a generic feature extractor.

\emph{Knowledge Transfer}~\cite{sindagi2020learning, zhao2020active}:
Sindagi et al.~\cite{sindagi2020learning} introduce a Gaussian Process (GP) to generate pseudo labels of the unlabeled data.
Zhao et al.~\cite{zhao2020active} propose to transfer feature representations across labeled and unlabeled data by a distribution classifier with the mixup technique.

\emph{Temporal Redundancy}~\cite{tan2011semi, loy2013semi, zhou2018crowd}:
Tan et al.~\cite{tan2011semi} propose a Semi-Supervised Elastic Net (SSEN) to regularize temporally neighboring samples.
Loy et al.~\cite{loy2013semi} analyze the geometric structure of crowd patterns and design the distribution and temporal regularization for manifold learning.
Zhou et al.~\cite{zhou2018crowd} propose a submodular method to annotate informative frames in crowd videos and introduce the graph Laplacian regularization for semi-supervised learning.

Despite the promising results of the above methods, they all adopt a None-or-All labeling strategy which inevitably introduces lots of labeling redundancy and lack diversity.
Differently, we propose to annotate the representative regions only in each crowd image and transfer the labeling redundancy caused by individual similarities to effective supervision signals on the unlabeled regions.

\section{Method}
\subsection{Framework Overview}

\begin{figure*}[t]
 \centering
 \setlength{\tabcolsep}{1pt}
 \includegraphics[width=.8\linewidth,height=7cm]{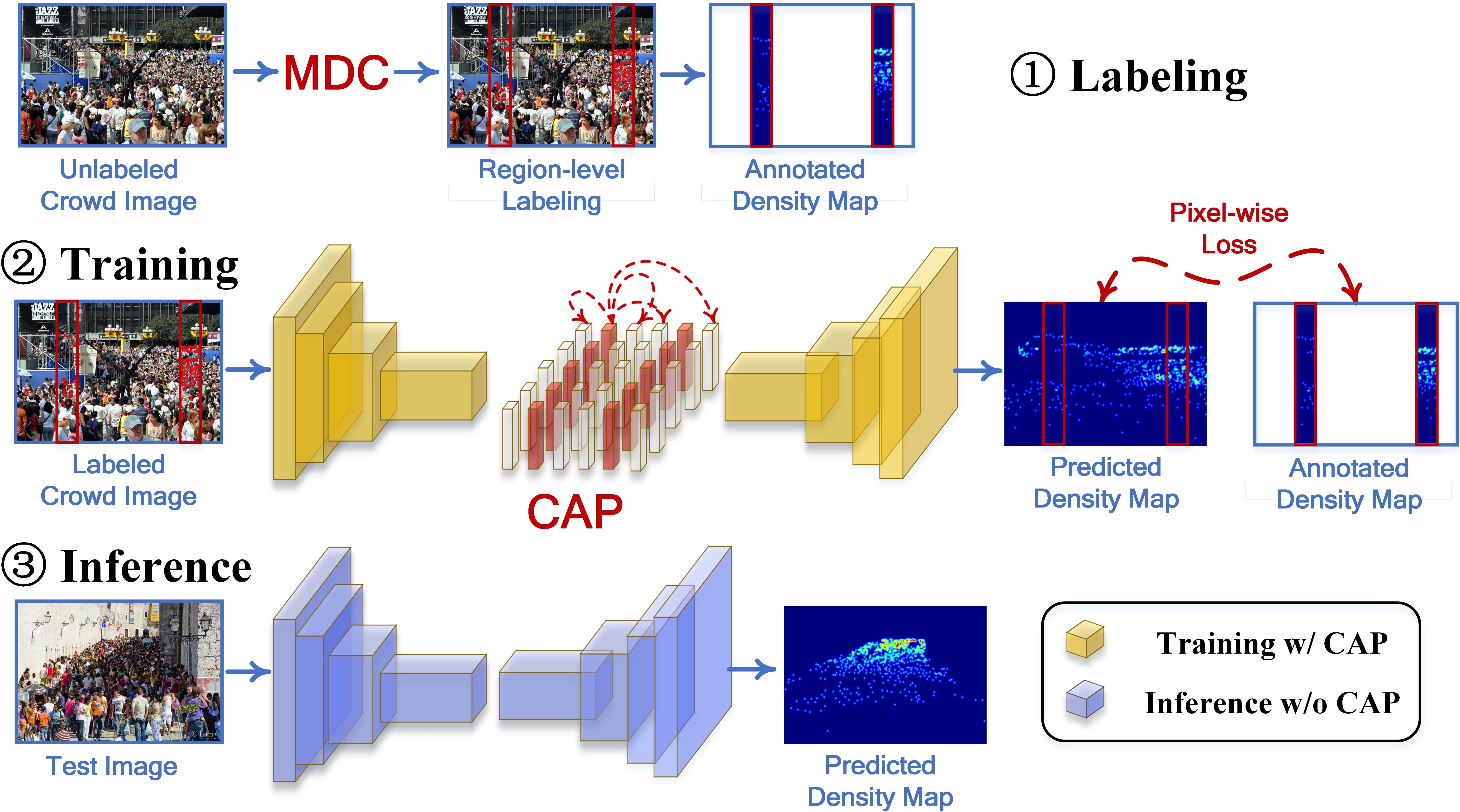}
 \caption{Overview of the proposed semi-supervised crowd counting framework, which consists of three stages, i.e., labeling, training, and inference. At the labeling stage, we design a Multi-level Density-aware Cluster (MDC) strategy to annotate the representative regions only in each crowd image. After labeling, to leverage the rich unlabeled regions, a Crowd Affinity Propagation (CAP) module is introduced to supervise both the labeled and unlabeled regions via feature propagation by exploiting the deep feature affinities among individuals.
 Note that the CAP module can be removed at the inference stage without performance degradation, which makes it computationally free after training.}
 \label{network architecture}
 \end{figure*}

As shown in Fig.~\ref{network architecture}, we propose a novel semi-supervised framework for crowd counting, which contains three major stages, i.e., labeling, training, and inference.

\subsection{Problem Formulation}

In semi-supervised crowd counting, we are given a limited labeling budget (e.g., 10\% of the training set).
After labeling, we have access to a labeled set which is denoted as $\mathcal{S_L} = \{(\mathbf{x}_{i}^{l}, \mathbf{y}_{i}^{l})\}_{i=1}^{N_{l}}$, where $\mathbf{x}_{i}^{l}$ and $\mathbf{y}_{i}^{l}$ denote the $i$-th annotated crowd image and its corresponding label, i.e., a set of coordinates pointing out the positions of head centers.
Besides, the remaining unlabeled samples form an unlabeled set $\mathcal{S_U}=\{(\mathbf{x}_{j}^{u})\}_{j=1}^{N_{u}}$.
Our goal is to utilize both sets to advance the crowd counting performance.
Note that different from previous methods, we do not label all the regions in each crowd image, so $\mathbf{x}_{i}^{l}$ and $\mathbf{x}_{j}^{u}$ are regions of crowd images in our context.

\subsection{Crowd Counting Network}
Crowd counting networks typically employ density maps as the intermediate output, which can be generated by convolving annotated head points with Gaussian kernels~\cite{zhang2016single}:
\begin{equation}
\label{density map generation}
    \mathcal{D}(\mathbf{z})=\sum_{k=1}^{N}\delta(\mathbf{z}-\mathbf{z}_{k}) * G_{\sigma_{k}}(\mathbf{z}),
\end{equation}
where $\mathbf{z}$ denotes each pixel in a crowd image $\mathbf{x}$. $\mathbf{z}_{k}$ represents the $k$-th annotated point (total $N$ points). $G_{\sigma_{k}}$ is a 2D Gaussian kernel with a bandwidth $\sigma_{k}$.
Therefore, the crowd counting problem is converted to: $\mathcal{F} : \mathcal{I}(\mathbf{x}) \rightarrow \mathcal{D}(\mathbf{x})$, which learns a mapping from an image space $\mathcal{I}(\mathbf{x})$ to a density map space $\mathcal{D}(\mathbf{x})$.
Following previous works~\cite{liu2020semi, zhao2020active, sindagi2020learning}, we employ a general and effective $\mathcal{F}$ based on CSRNet~\cite{li2018csrnet} to evaluate the effectiveness of proposed semi-supervised methods.
To train $\mathcal{F}$, we adopt the pixel-wise Euclidean loss
to measure the distance between the annotated and estimated density maps:
\begin{equation}
\label{counting loss function}
    \mathcal{L}_{den}(\Theta) = \frac{1}{2M}\sum_{m=1}^{M} \left\| \mathcal{F}(\mathcal{I}_m; \Theta) - \mathcal{D}_m \right\|_{2}^{2},
\end{equation}
where $\Theta$ is the learnable parameters of $\mathcal{F}$. $\mathcal{I}_m$ is the $m$-th training image (total $M$ images). $\mathcal{F}(\mathcal{I}_m; \Theta)$ and $\mathcal{D}_m$ denote the estimated and annotated density maps, respectively.

\subsection{Representative Regions Selection Strategy}
\label{RRSS}
As we want to label more crowd images with diverse individuals, we transfer the labeling budget to each crowd image.
For example, if the budget is 10\% of the entire dataset, we choose to label all the crowd images with 10\% of each annotated.
Then we come to the problem of how to find the representative regions in each crowd image.

\noindent \textbf{Annotate More in the Vertical or Horizontal Direction?}
Regions in crowd images can be categorized into three types: dense , sparse, and background regions according to different crowd distributions.
The representative regions in each crowd image should cover all the three types and have as large crowd density variations as possible given a limited labeling budget.
As shown in Fig.~\ref{teaser}, large crowd density variations usually appear in the vertical direction (e.g., from bottom to top) of each crowd image due to the surveillance camera perspective and imaging condition.

Therefore, when we are given a labeling budget in each crowd image, we should label a region which spreads more in the vertical direction than the horizontal direction (see Table~\ref{vertical or horizontal} for an experimental comparison).
Without loss of generality, we define the labeled region $\mathbf{x}_l \in \mathbb{R}^{H_l \times W_l}$ as a rectangular region of a crowd image $\mathbf{x} \in \mathbb{R}^{H \times W}$.
As discussed above, $H_l$ should be much larger than $W_l$ (i.e., $H_l \gg W_l$).
In the extreme case, $H_l = H$ and $W_l$ varies according to the labeling budget.
In practice, $\mathbf{x}_l$ may not be a continuous area in $\mathbf{x}$ and may contain $n_l$ subregions.
Therefore, the representative regions selection problem is simplified to determine the $n_l$ subregions in $\mathbf{x}$.

\noindent \textbf{Multi-level Density-aware Cluster (MDC) Strategy.}
 \begin{figure}[t]
 \centering
 \setlength{\tabcolsep}{1pt}
 \includegraphics[width=.95\linewidth]{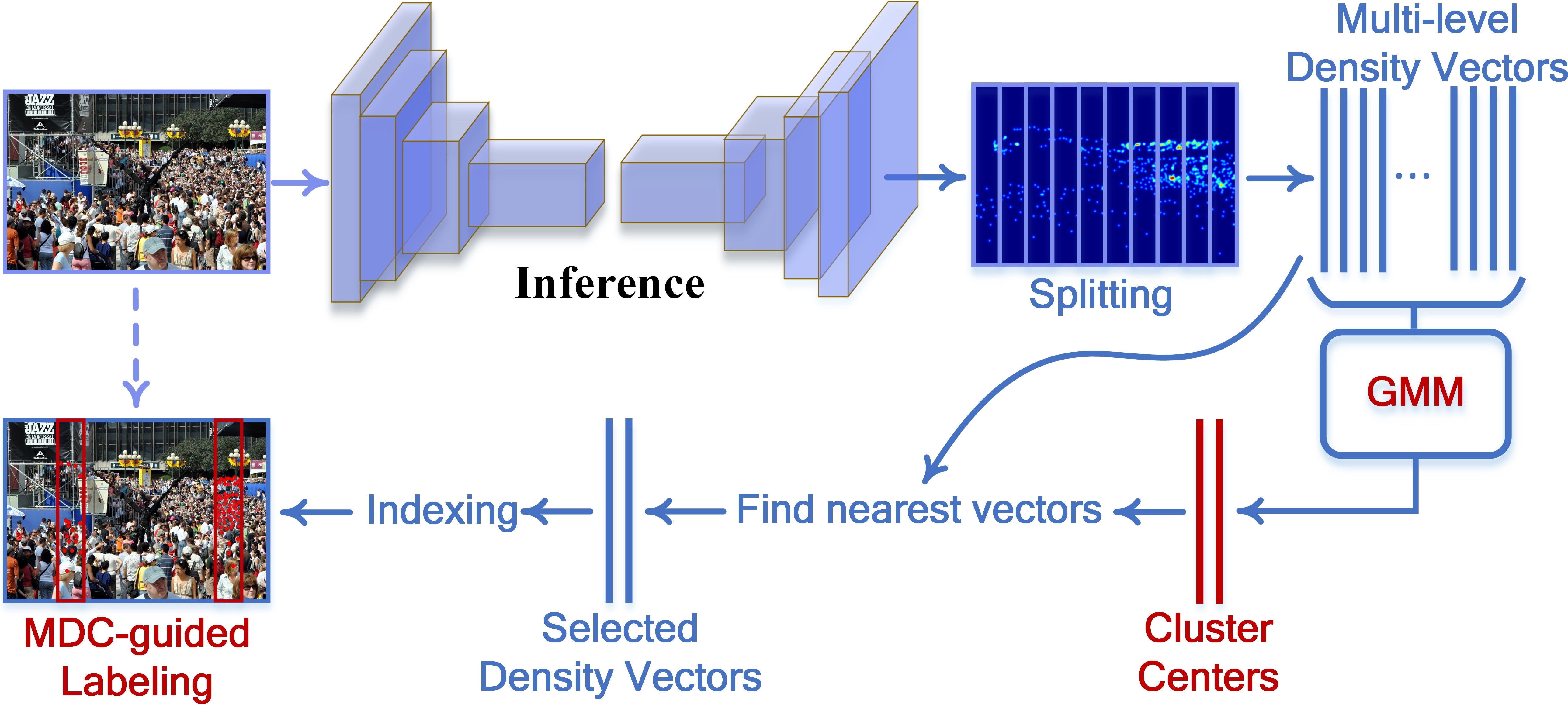}
 \caption{Illustration of the proposed Multi-level Density-aware Cluster strategy for representative regions selection.}
 \label{MDC figure}
 \end{figure}
The MDC strategy shown in Fig.~\ref{MDC figure} is designed to determine the $n_l$ subregions in $\mathbf{x}$.
First, we divide $\mathbf{x}$ into $n_u$ rectangular subregions $\mathbf{x}_u = \{\mathbf{x}_u^1, \mathbf{x}_u^2, ..., \mathbf{x}_u^{n_u}\}$ with each $\mathbf{x}_u^j \in \mathbb{R}^{H \times W_u^j}$ and $W_u^j = Constant$.
Then, our goal is to select $n_l$ subregions $\mathbf{x}_l = \{\mathbf{x}_l^1, \mathbf{x}_l^2, ..., \mathbf{x}_l^{n_l}\}$ from $\mathbf{x}_u$.
Different from the crowd density variations in the vertical direction, the crowd scene usually changes in the horizontal direction due to the large-view field of cameras.
Therefore, we should select as many crowd scenes as possible in the horizontal direction.
Based on the definition that the same crowd scene shares the same crowd density distributions along the horizontal direction, we propose a Multi-level Density-aware Cluster strategy to cluster the unlabeled regions $\mathbf{x}_u$ into multiple crowd scenes based on their multi-level density vectors.

Specifically, to obtain density distributions of the unlabeled data,
we first randomly label a few crowd images (e.g., 20\% of the labeling budget) as warm-up samples to pretrain a crowd counter.
For each unlabeled region $\mathbf{x}_u^j$ in a crowd image $\mathbf{x}$, we extract its predicted density maps $\mathbf{m}_u^j$ and calculate the multi-level density vector $\mathbf{v}_u^j$ of $\mathbf{m}_u^j$ as:
\begin{equation}
\label{multi-scale density vector}
    \mathbf{v}_u^j =[\mathcal{V}^1(\mathbf{m}_u^j), \mathcal{V}^2(\mathbf{m}_u^j), ..., \mathcal{V}^L(\mathbf{m}_u^j)],
\end{equation}
where $\mathcal{V}^L(\mathbf{m}_u^j)$ is the $L$-th level density vector defined as:
\begin{equation}
\label{each density vector}
\begin{split}
    \mathcal{V}^L(\mathbf{m}_u^j) =[&\,Sum(\mathbf{m}_u^j[H_L\!:\!H_L*1, :]), \\
                                    &\,Sum(\mathbf{m}_u^j[H_L*1\!:\!H_L*2, :]),\\
                                    &\,..., \\
                                    &\,Sum(\mathbf{m}_u^j[H_L*(L-1)\!:\!H_L*L, :])\,],
\end{split}
\end{equation}
where $Sum(\cdot)$ and $*$ denote the summation and multiplication operations, respectively.
$H_L$ is equal to $H$ integrally divided by $L$.
$\mathbf{m}_u^j[:, :]$ means a subregion of $\mathbf{m}_u^j$ where the former and latter dimensions are height and width, respectively.
As the initial values in $\mathbf{v}_u^j$ have different scales, we normalize each of them to the same scale by $\mathcal{V}^k(\mathbf{m}_u^j)/L*k$ where $k$ and $/$ denotes the $k$-th level and the division operation, respectively.
With the multi-level design, $\mathbf{v}_u^j$ can express both the local and global crowd density distributions of $\mathbf{d}_u^j$.
 
Based on the calculated multi-level density vectors, we introduce a probabilistic cluster algorithm based on Gaussian Mixture Models (GMM) to cluster the unlabeled regions $\mathbf{x}_u$ into multiple crowd scenes.
In particular, we initialize a Gaussian mixture distribution as follows:
\begin{equation}
\label{gaussian mixture distribution}
    p_{\mathcal{M}}(\mathbf{v}_u) = \sum_{n=1}^{n_s} \alpha_n \cdot p(\mathbf{v}_u | \mathbf{\mu}_n, \mathbf{\Sigma}_n),
\end{equation}
where $n_s$ is the number of multivariate Gaussian distributions. $\alpha_n$ denotes the mixture coefficient and $\sum_{n=1}^{n_s}\alpha_n=1$. $p(\mathbf{v}_u | \mathbf{\mu}_n, \mathbf{\Sigma}_n)$ is the $n$-th multivariate Gaussian distribution where $\mathbf{\mu}_n$ and $\mathbf{\Sigma}_n$ are the mean and covariance, respectively.
To resolve the parameters ${(\alpha_n, \mathbf{\mu}_n, \mathbf{\Sigma}_n} | 1 \leq n \leq n_s)$ given $\mathbf{v}_u$, we adopt Maximum Likelihood Estimation (MLE) to maximize the following log-likelihood:
\begin{equation}
\label{maximum likelihood estimation}
\begin{split}
    LL(\mathbf{v}_u) &= ln \left(\prod_{j=1}^{n_u} p_{\mathcal{M}}(\mathbf{v}_u^j)\right)\\
    &=\sum_{j=1}^{n_u}ln\left(\sum_{n=1}^{n_s}\alpha_n \cdot p(\mathbf{v}_u^j | \mathbf{\mu}_n, \mathbf{\Sigma}_n) \right).
\end{split}
\end{equation}
Following previous methods, we utilize the EM algorithm~\cite{mclachlan2007algorithm} to iteratively optimize the parameters.
When the optimization process converges, we can obtain the posterior distribution of $\mathbf{v}_u^j$ according to the Bayes' theorem:
\begin{equation}
\label{get componet}
\begin{split}
    p_{\mathcal{M}}(z_j = n | \mathbf{v}_u^j) &= \frac{P(z_j = n)\cdot p_{\mathcal{M}}(\mathbf{v}_u^j | z_j = n)}{p_{\mathcal{M}}(\mathbf{v}_u^j)}\\
    &= \frac{\alpha_n \cdot p(\mathbf{v}_u^j | \mathbf{\mu}_n, \mathbf{\Sigma}_n)}{\sum_{r=1}^{n_s}\alpha_r  \cdot p(\mathbf{v}_u^j | \mathbf{\mu}_r, \mathbf{\Sigma}_r)},
\end{split}
\end{equation}
where $z_j \in \{1,2,...,n_s\}$ is each component of the Gaussian mixture distribution.
The cluster tag $\lambda_u^j$ of each multi-level density vector $\mathbf{v}_u^j$ can be determined by $\mathop{\arg\max}_{n\in{1,2,...,n_s}} p_{\mathcal{M}}(z_j = n | \mathbf{v}_u^j)$.
For all the density vectors with the same cluster tag, we calculate the Euclidean distance between each vector and the mean of the corresponding Gaussian model and select the vector with the smallest distance to act as the representative region of each cluster.
After obtaining the $n_l$ representative regions, we merge the spatially adjacent ones and label them to generate the annotated density maps.

\subsection{Crowd Affinity Propagation}
To leverage the rich unlabeled regions, we propose a novel Crowd Affinity Propagation (CAP) module to directly supervise the unlabeled regions via feature propagation by exploiting deep feature affinities.
The rationale behind this design is that prediction by comparison is more effective than direct prediction for the cases with only limited annotations.
Crowd counting typically requires sufficient supervision to capture the diverse data distributions.
However, the affinities between deep features in the latent space can help infer whether they belong to the same class via relatively low-level semantics, e.g., similar color and texture.

Specifically, the CAP module contains two phases, e.g., forward propagation and backward propagation.
In the forward propagation, deep features from the unlabeled regions are transferred to update those of the labeled regions by leveraging the feature affinities between them.
Let $\mathbf{f}_u \in \mathbb{R}^{C\times H_u \times W_u}$ and $\mathbf{f}_l \in \mathbb{R}^{C\times H_l \times W_l}$ denote the features extracted by the crowd counter from the unlabeled and labeled regions in a crowd image. As $\mathbf{f}_u$ and $\mathbf{f}_l$ are extracted synchronously, they share the same number of channel dimensions $C$.
The initial values in $\mathbf{f}_u$ and $\mathbf{f}_l$ may be very large or small, so we first normalize them as follows:
\begin{equation}
\label{normalize feature}
\mathbf{f}_u^n = \mathcal{S}(\mathbf{f}_u + eps), \mathbf{f}_l^n = \mathcal{S}(\mathbf{f}_l + eps),
\end{equation}
where $\mathbf{f}_u^n$ and $\mathbf{f}_l^n$ are the normalized features. $\mathcal{S}(\cdot)$ denotes the softmax function along the channel dimension. $eps$ is a small value to avoid dividing by zero.
For the sake of clarity, we still utilize $\mathbf{f}_u$ and $\mathbf{f}_l$ instead of $\mathbf{f}_u^n$ and $\mathbf{f}_l^n$ to denote the normalized features.
We reshape $\mathbf{f}_u$ to $\mathbb{R}^{C\times N_u}$ where $N_u = H_u \times W_u$ and $\mathbf{f}_l$ to $\mathbb{R}^{C\times N_l}$ where $N_l = H_l \times W_l$, and then $\mathbf{f}_u=\{\mathbf{f}_u^1, \mathbf{f}_u^2, ..., \mathbf{f}_u^{N_u}\}$ and $\mathbf{f}_l=\{\mathbf{f}_l^1, \mathbf{f}_l^2, ..., \mathbf{f}^{N_l}\}$ with each feature in $\mathbf{f}_u$ or $\mathbf{f}_l$ has $C$ dimensions.
Then we calculate the normalized similarity $\mathbf{s}_{ij}$ between each feature $\mathbf{f}_l^i$ of $\mathbf{f}_l$ and each feature $\mathbf{f}_u^j$ of $\mathbf{f}_u$ as follows:
\begin{equation}
\label{feature similarity}
\mathbf{s}_{ij} = \frac{exp(\mathbf{f}_l^i \cdot \mathbf{f}_u^j)}{\sum_{k=1}^{N_u}exp(\mathbf{f}_l^i \cdot \mathbf{f}_u^k))}.
\end{equation}
The more similar $\mathbf{f}_l^i$ and $\mathbf{f}_u^j$ are, the higher $\mathbf{s}_{ij}$.
After calculating the feature similarities, we propagate all the features of the unlabeled regions $\mathbf{f}_u$ to update each feature $\mathbf{f}_l^i$ of $\mathbf{f}_l$, which can be defined as:
\begin{equation}
\label{feature propagate}
\mathbf{f}_l^{i+} = \gamma \cdot \sum_{j=1}^{N_u}\mathbf{s}_{ij} \cdot \mathbf{f}_u^j + (1 - \gamma) \cdot \mathbf{f}_l^i,
\end{equation}
where $\mathbf{f}_l^{i+}$ is the updated $i$-th feature $\mathbf{f}_l^i$ of $\mathbf{f}_l$. $\gamma$ is a learnable parameter to fuse the labeled and unlabeled features.
After feature updating, the features of the labeled regions can also contain those of the unlabeled regions, which can be supervised by the labels of the labeled regions in the backward propagation.
When the training procedure converges, we can optionally remove the CAP module from the crowd counter without performance degradation (see Table~\ref{CAP TABLE} for a detailed comparison) which means the proposed CAP module is computationally free at the inference stage.

\begin{algorithm}[t]
\caption{The Proposed Semi-supervised Crowd Counting Framework.}
\label{optimization procedure}
\LinesNumbered 
\KwIn{An unlabeled dataset $\mathcal{D}$. \qquad \qquad \qquad \qquad
      Labeling budget $M\%$. \qquad \qquad \qquad \qquad \qquad
      Percentage of warm-up samples $R\%$.}
\KwOut{A crowd counter $\mathcal{F}$}
Randomly label $|\mathcal{D}|*R\%$ samples of $\mathcal{D}$ to form $\mathcal{D}_{pre}$ with $M\%$ regions of each sample annotated. \\
Pretrain $\mathcal{F}$ in $\mathcal{D}_{pre}$ by Eq.~(\ref{counting loss function}). \\
Obtain density maps $\mathbf{m}_u$ of the remaining unlabeled samples (i.e., $\mathcal{D}_u = \mathcal{D} - \mathcal{D}_{pre}$) by $\mathcal{F}$. \\
Calculate multi-level density vectors $\mathbf{v}_u$ of $\mathbf{m}_u$ according to Eq.~(\ref{multi-scale density vector}) and Eq.~(\ref{each density vector}). \\
Cluster to select representative regions $\mathbf{x}_l$ in each crowd image $\mathbf{x}$ of $\mathcal{D}_u$ according to Eq.~(\ref{each density vector}). \\
Label representative regions $\mathbf{x}_l$. \\
Train $\mathcal{F}$ with CAP in $\mathcal{D}_{pre} + \mathcal{D}_{u}$. \\
Infer $\mathcal{F}$ without CAP.
\end{algorithm}

\subsection{Network Optimization}
The proposed semi-supervised crowd counting framework is shown in Algorithm~\ref{optimization procedure}, which contains three stages: (i) labeling, (ii) training with CAP, and (iii) inference without CAP. At the labeling stage, we first randomly label a small portion of the labeling budget as warm-up samples to pretrain a crowd counter for multi-level density vectors utilized in the MDC strategy.
Then we label the remaining samples by the MDC strategy.
After labeling, we train the crowd counter with the CAP module by all the labeled samples.
At the inference stage, we remove the CAP module from the crowd counter and estimate density maps and crowd counts for any given crowd images.

\section{Experiments}

\subsection{Experimental Setup}
\noindent \textbf{Datasets.}
Three widely-used crowd counting datasets are employed in our experiments.
(i) \emph{ShanghaiTec PartA} \cite{zhang2016single} is collected from the Internet containing 482 images, in which 300 images for training and the remaining for testing.
(ii) \emph{ShanghaiTec PartB} \cite{zhang2016single} is collected from the metropolitan areas in Shanghai consisting of 716 images, in which 400 images for training and the remaining for testing.
Compared to PartA, PartB has relatively fixed crowd scenes and less crowd counts.
(iii) \emph{UCF-QNRF} \cite{idrees2018composition} is a very challenging dataset, which contains 1,535 crowd images with 1,251,642 annotations.
The training set consists of 1,201 images, while the testing set has 334 images.
Compared to the above two datasets, UCF-QNRF is bigger and contains more various crowd distributions as well as higher image resolutions.

\noindent \textbf{Implementation Details.}
As image resolutions in crowd counting datasets vary greatly, we set the batch size as 1 in all experiments.
The width $W_u^j$ of each unlabeled subregion $\mathbf{x}_u^j$ in Sec.~\ref{RRSS} should not be too large or too small.
If it is too large, the proposed region-level strategy may degrades into the None-or-All strategy.
If too small, we cannot annotate human heads in each subregion with very few visual cues.
We empirically set it as 10\% of the width of the corresponding crowd image in all datasets.
This means that each subregion takes up 10\% of the entire crowd image.
We utlize random cropping and horizontal flipping for data augmentation.
$\sigma_{k}$ in Eq.~(\ref{density map generation}) is a fixed bandwidth and is set as 4 for all datasets.
$L$ in Eq.~(\ref{multi-scale density vector}) is set as 4 via cross validation to balance the efficiency and computational cost.
$eps$ in Eq.~(\ref{normalize feature}) is set as $\mathrm{10}^{-\mathrm{6}}$ and $\gamma$ in Eq.~(\ref{feature propagate}) is initialized as 0.2.
Adam optimizer~\cite{kingma2014adam} is employed to optimize the crowd counting network with the initial learning rate as $\mathrm{10}^{-\mathrm{8}}$.
The experiments are conducted on NVIDIA GTX 2080Ti GPU.

\noindent \textbf{Evaluation protocol.}
Following previous methods~\cite{zhang2016single}, Mean Absolute Error (MAE) and Root Mean Squared Error
(RMSE) are used as the evaluation metrics.
  \begin{equation}\label{MAE RMSE}
    MAE = \frac{1}{S} \sum_{i=1}^{S} | \hat{C}_i - C_i^{GT} |,\ RMSE = \sqrt{\frac{1}{S} \sum_{i=1}^{S} | \hat{C}_i - C_i^{GT} |^2},
  \end{equation}
  where $S$ is the total number of testing samples. $\hat{C}_i$ and $C_i^{GT}$ are the estimated and ground-truth crowd counts of the $i$-th sample which are calculated by summing over the entire density maps.
  As the random sample selection and network training have uncertainty, we train and test each method for 5 times and calculate the mean of MAE and RMSE to evaluate the performance of each variant method.
  
  \begin{figure}[t]
 \centering
 \setlength{\tabcolsep}{1pt}
 \includegraphics[width=\linewidth,height=5cm]{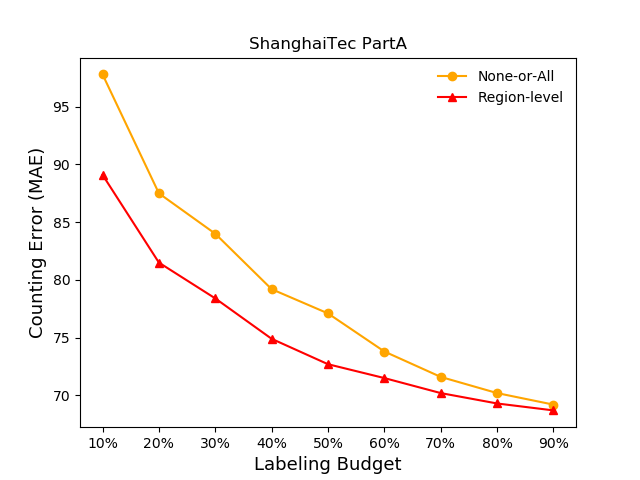}
 \caption{Comparison results of the None-or-All and region-level labeling strategies with respect to different labeling budgets in the ShanghaiTec PartA dataset.}
 \label{imageORregion}
 \end{figure}

\subsection{Ablation Studies}
We conduct extensive ablation studies in the ShanghaiTec PartA dataset \cite{zhang2016single} to validate the effectiveness of the proposed semi-supervised crowd counting method.

\noindent \textbf{Is region-level really better than None-or-All?}
To validate the effectiveness of the region-level labeling strategy, we randomly label images (for None-or-All strategy) and regions (for region-level strategy) and compare their performance with respect to different labeling budgets (i.e., 10\%, 20\%, 50\%, 70\% and 90\%).
The comparison results are shown in Fig~\ref{imageORregion}.
We can see that the region-level strategy can consistently outperform the None-or-All strategy.
Note that the performance gain of the region-level strategy is magnified when the labeling budget gets small.
This indicates that it is better to employ the region-level labeling strategy to annotate more crowd images with various crowd scenes, especially when the labeling budget is limited.

\begin{table}[t]
	\centering
	\small
	\tabcolsep=0.15cm
	\caption{Comparison results of different spatial ratios of the labeled regions in each crowd image. The experiments are conducted in the ShanghaiTec PartA dataset with a 10\% labeling budget. *:* denotes vertical:horizontal. $\infty$:1 (or 1:$\infty$) represents the spatial ratio with the height (or width) of the labeled region equal to that of the entire image.}
	\begin{tabular}{c||ccccccc}
	    \toprule
        Ratio&1:$\infty$&1:4&1:2&1:1&2:1&4:1&$\infty$:1 (ours) \cr
        \midrule
        \midrule
        MAE &95.9& 95.2 & 93.8 & 93.3& 92.4& 91.3&\textbf{89.1}\cr
        RMSE &148.3& 145.8 & 144.5 & 143.8& 142.0& 140.5&\textbf{137.5}\cr
        \bottomrule
        \end{tabular}
	\label{vertical or horizontal}
\end{table}

\noindent \textbf{Annotate more in the vertical or horizontal direction?}
To validate the effectiveness of the vertical-first annotation strategy, we fix the labeling budget~(i.e., 10\%) in each crowd image and change the spatial ratios (i.e., vertical:horizontal) of the randomly labeled regions.
The comparison results are shown in Table~\ref{vertical or horizontal}.
We can see that with more annotations in the vertical direction (e.g., the extreme case is $\infty$:1 where the height of the labeled region is equal to that of the entire image), the counting performance can be enhanced gradually, which confirms the effectiveness of the vertical-first annotation strategy considering the large crowd density variations caused by the camera perspectives.

\begin{table}[t]
	\centering
	\small
	\tabcolsep=0.15cm
	\caption{Comparison results of different representative regions selection strategies in the ShanghaiTec PartA dataset. 
	Each result is in the form of MAE/RMSE.}
	\begin{tabular}{c||cccc}
	    \toprule
        Method&10\%&20\%&50\%&90\%\cr
        \midrule
        \midrule
        RANDOM & 89.1/137.5 & 81.5/128.4 & 72.6/120.5& 68.7/116.8\cr
        MAX & 90.8/133.3 & 82.6/126.7  & 71.9/119.1 & 69.1/\textbf{116.3}\cr
        MDC  &\textbf{83.3}/\textbf{132.1} & \textbf{76.4}/\textbf{125.2} & \textbf{71.2}/\textbf{118.4}& \textbf{68.5}/116.6\cr
        \bottomrule
        \end{tabular}
	\label{MDC table}
\end{table}

\begin{table}[t]
	\centering
	\small
	\tabcolsep=0.15cm
	\caption{Comparison results of different percentages of the warm-up samples (i.e., randomly labeled samples) utilized in the MDC strategy with a 10\% labeling budget in the ShanghaiTec PartA dataset.}
	\begin{tabular}{c||cccc}
	    \toprule
        Percentage of Warm-up Samples&10\%&20\%&30\%&40\%\cr
        \midrule
        \midrule
        MAE & 77.1 & \textbf{76.4} & 76.8& 78.5\cr
        RMSE & 125.8 & \textbf{125.2} & 126.3 &127.9\cr
        \bottomrule
        \end{tabular}
	\label{Percentage of Warm-up Samples}
\end{table}

\noindent \textbf{Effectiveness of the MDC strategy for representative regions selection.}
Based on the above two ablation studies, we have demonstrated the efficacy of the proposed region-level and vertical-first labeling strategies.
In this part, we further evaluate the effectiveness of the proposed MDC strategy for representative regions selection.
The experiments are conducted with respect to different region selection strategies (i.e., RANDOM, MAX, and MDC) where MAX means to select the regions with the maximum numbers of people in each crowd image.
Except RANDOM, both MAX and MDC need to warm up the crowd counter by randomly labeling some samples to generate crowd density maps of the unlabeled images.
We fix the percentage of the randomly labeled samples (namely warm-up samples), i.e., 20\% of the labeling budget and vary the labeling budgets to compare the different region selection strategies.

The comparison results are shown in Table~\ref{MDC table}.
We can see that the MDC strategy outperforms the other two strategies consistently when the labeling budget is limited (e.g., 10\%, 20\%, and 50\%).
When the labeling budget is abundant (e.g., 90\%), the counting performance of the three labeling strategies is saturated without obvious differences.
Note that the MDC strategy annotates less than the MAX strategy, e.g., when the labeling budget is 10\%, MDC annotates 15,939 human heads while MAX annotates 20,401.
This indicates that the proposed MDC strategy can achieve superior performance with less annotation burden, which demonstrates the effectiveness of the proposed multi-level density-aware cluster strategy for representative regions selection.

Besides, we explore the optimal percentage of the warm-up samples used in MDC by fixing the labeling budget (i.e., 10\%) and changing it from 10\% to 40\%.
The comparison results are summarized in Table~\ref{Percentage of Warm-up Samples}. We can see that 20\% is the optimal setting with the balance of a well-pretrained crowd counter and a large number of unlabeled samples for the representative regions selection.

\noindent \textbf{Effectiveness of the CAP module.}
To evaluate the effectiveness of the CAP module, we add the proposed CAP module based on the well-performed MDC module.
Two variants are designed to explore the optimal setting of the CAP module, which are denoted as ``CAP (train\&infer)'' and ``CAP (train only)'' in Table~\ref{CAP TABLE}.
``CAP (train\&infer)'' means to add the CAP module both at the training and inference stages, while ``CAP (train only)'' removes the CAP module after the training stage.
The comparison results are shown in Table~\ref{CAP TABLE}.
We can see that with the CAP module, the counting performance of both ``MDC + CAP (train\&infer)'' and ``MDC + CAP (train only)'' can be improved considerably in multiple labeling budgets, which demonstrates the effectiveness of exploiting deep feature affinities to directly supervise the unlabeled regions.
Besides, by the comparison between ``MDC + CAP (train\&infer)'' and ``MDC + CAP (train only)'', we find that the CAP module can be removed at the inference stage without performance degradation. This indicates that the proposed CAP module can enhance the counting performance efficiently without extra computational costs after training.

\begin{figure}[t]
 \begin{center}
 \setlength{\tabcolsep}{1.0pt}
 \renewcommand{\arraystretch}{0.6}
  \begin{tabular}{ccc}
     \includegraphics[width=.29\linewidth, height=1.6cm]{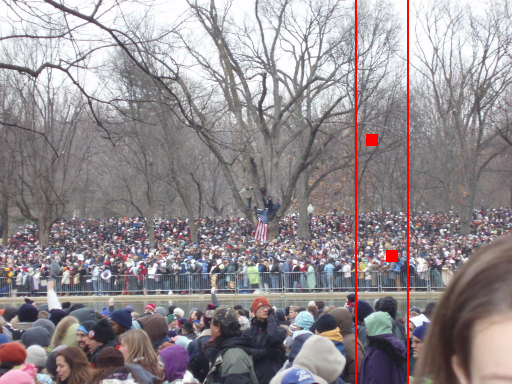}&
     \includegraphics[width=.29\linewidth, height=1.6cm]{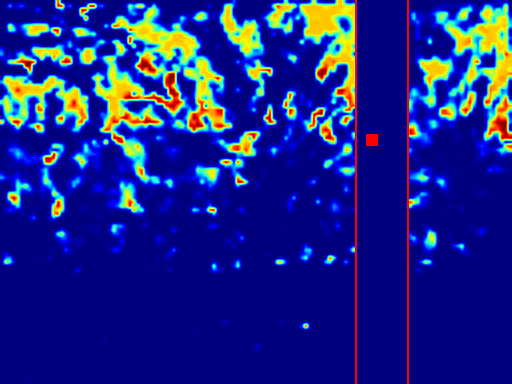}&
     \includegraphics[width=.29\linewidth, height=1.6cm]{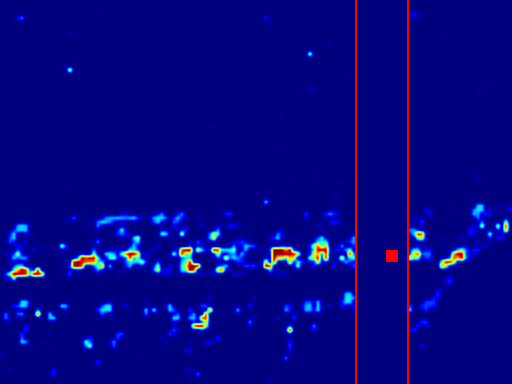}\\
     \includegraphics[width=.29\linewidth, height=1.6cm]{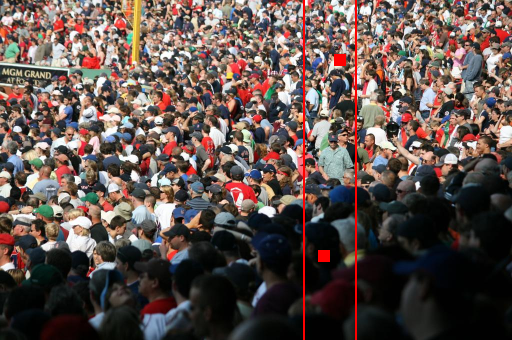}&
     \includegraphics[width=.29\linewidth, height=1.6cm]{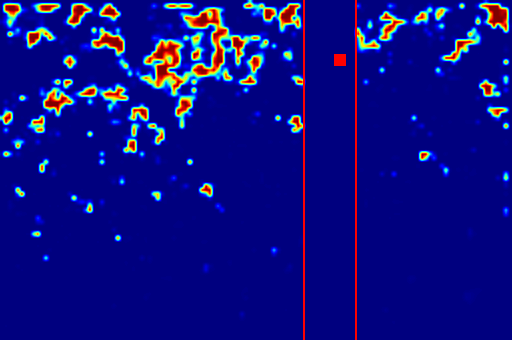}&
     \includegraphics[width=.29\linewidth, height=1.6cm]{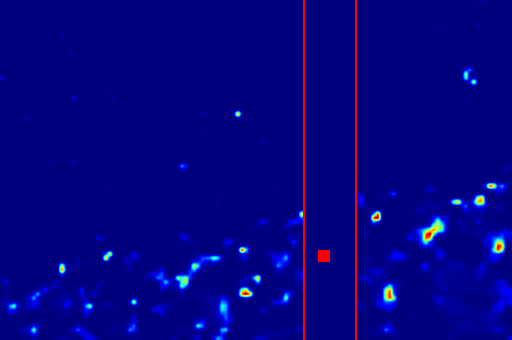}\\
     \includegraphics[width=.29\linewidth, height=1.6cm]{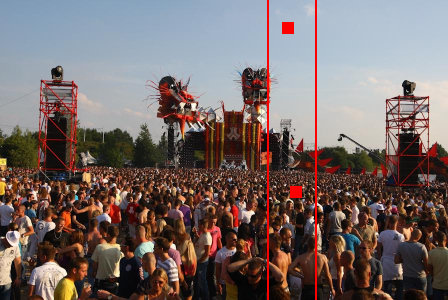}&
     \includegraphics[width=.29\linewidth, height=1.6cm]{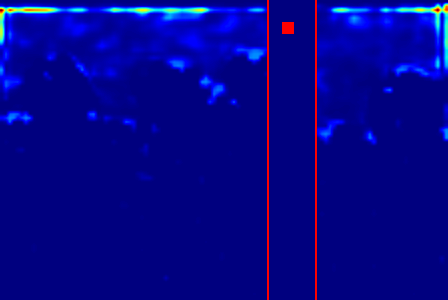}&
     \includegraphics[width=.29\linewidth, height=1.6cm]{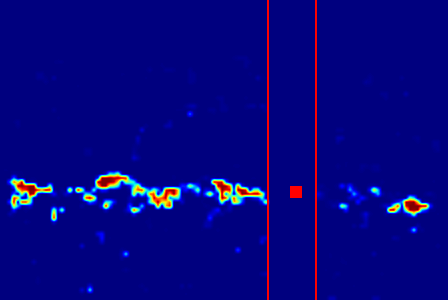}\\
     Input Image & Affinity Map $\#1$ & Affinity Map $\#2$ \\
  \end{tabular}
\end{center}
\vspace{-3pt}
\caption{Visualization of affinity maps in the CAP module. Labeled regions are enclosed between two red lines in each crowd image with two representative positions marked. Crowd affinities between each marked position and all the positions of unlabeled regions are illustrated in the latter two columns. Warmer colors mean higher values.}
\vspace{-3pt}
\label{CAP figure}
\end{figure}

\begin{table}[t]
	\centering
	\small
	\tabcolsep=0.15cm
	\caption{Ablation studies on the CAP module with respect to different labeling budgets in the the ShanghaiTec PartA dataset. Each result is in the form of MAE/RMSE.}
	\begin{tabular}{c||ccc}
	    \toprule
        Method&10\%&20\%&50\%\cr
        \midrule
        \midrule
        MDC & 83.3/132.1 & 76.4/125.2 & 71.2/118.4\cr
        MDC + CAP (train\&infer) & \textbf{78.8}/127.9 & \textbf{72.7}/\textbf{120.6} & \textbf{68.5}/116.1\cr
        MDC + CAP (train only) & 79.6/\textbf{127.5} & 73.2/121.3 & 69.2/\textbf{115.7}\cr
        \bottomrule
        \end{tabular}
	\label{CAP TABLE}
\end{table}

\begin{table}[t]
	\centering
	\small
	\tabcolsep=0.05cm
	\caption{Comparison results with state-of-the-art methods in the ShanghaiTec PartA \cite{zhang2016single} (denoted as STPart A), ShanghaiTec PartB \cite{zhang2016single} (denoted as STPart B), and UCF-QNRF \cite{idrees2018composition} datasets. The labeling budgets are 10\%, 10\%, and 20\%, respectively. ``S'' and ``F'' denote semi-supervised and fully-supervised methods, respectively.}
	\begin{tabular}{cc||cc|cc|cc}
		\toprule
		\multirow{2}{*}{Method}&\multirow{2}{*}{Type}&
		\multicolumn{2}{c|}{STPart A}&\multicolumn{2}{c|}{STPart B}&\multicolumn{2}{c}{UCF-QNRF}\\
		&&MAE$\downarrow$&RMSE$\downarrow$&MAE$\downarrow$&RMSE$\downarrow$&MAE$\downarrow$&RMSE$\downarrow$\\
        \midrule
        CSRNet \cite{li2018csrnet} & F & 68.2 & 115.0& 10.6 & 16.0 & 121.3 & 215.2\cr
        MT \cite{tarvainen2017mean} & S & 94.5 & 156.1 & 15.6 & 24.5 & 145.5 & 250.3\cr
        UDA \cite{xie2020unsupervised} & S & 93.8 & 157.2 & 15.7 & 24.1 & 144.7 & 255.9  \cr
        L2R \cite{liu2019exploiting} & S & 90.3 & 153.5 & 15.6 & 24.4 & 148.9 & 249.8\cr
        IRAST \cite{liu2020semi} & S & 86.9 & 148.9& 14.7 & 22.9 & 135.6 & 233.4\cr
        AL-AC \cite{zhao2020active} & S & 87.9 & 138.8& 13.9 & 26.2 & -- & -- \cr
        Ours & S & \textbf{79.6} & \textbf{127.5} & \textbf{12.7} & \textbf{20.3} & \textbf{128.6} & \textbf{226.4}\cr
        \bottomrule
        \end{tabular}
	\label{comparison with sota}
\end{table}

 \begin{figure}[!t]
 \centering
 \setlength{\tabcolsep}{0.6pt}
 \renewcommand{\arraystretch}{0.6}
  \begin{tabular}{ccccc}
     & \rotatebox[origin=l]{90}{\footnotesize{Input Image}}&
     \includegraphics[width=.31\linewidth]{./figures/QR/IMG_24_img.png}&
     \includegraphics[width=.31\linewidth]{./figures/QR/IMG_17_img.png}&
     \includegraphics[width=.31\linewidth]{./figures/QR/IMG_116_img.png}\\
     & \rotatebox[origin=l]{90}{\footnotesize{None-or-All}}&
     \includegraphics[width=.31\linewidth]{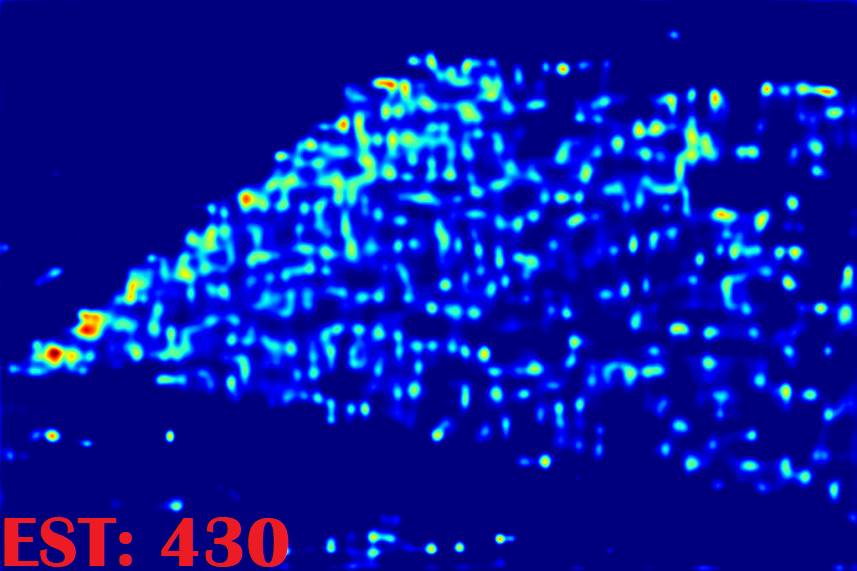}&
     \includegraphics[width=.31\linewidth]{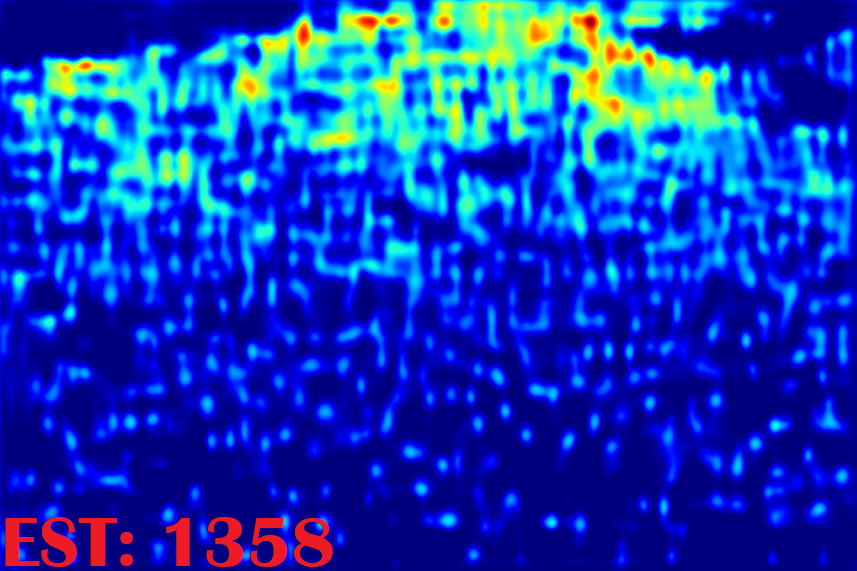}&
     \includegraphics[width=.31\linewidth]{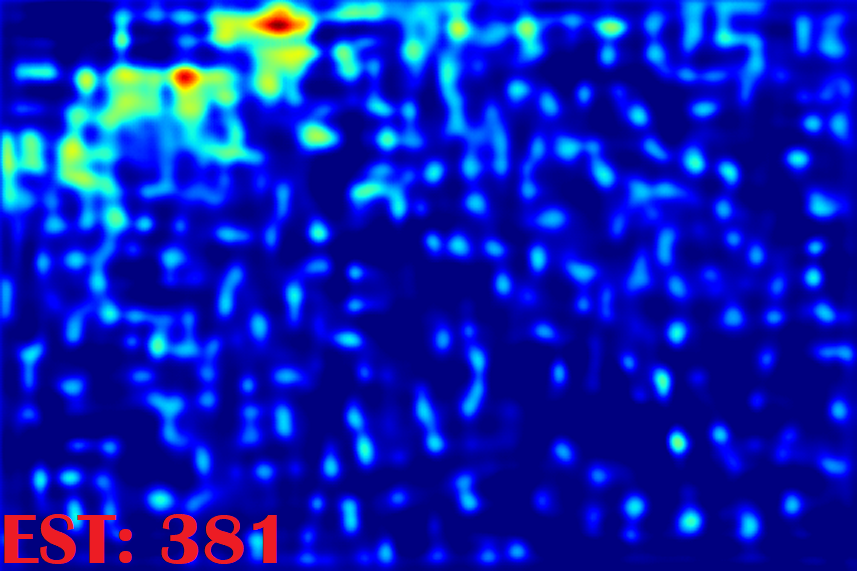}\\
     \rotatebox[origin=l]{90}{\hspace{6mm}\footnotesize{w/o}} & \rotatebox[origin=l]{90}{\footnotesize{MDC\&CAP}}&
     \includegraphics[width=.31\linewidth]{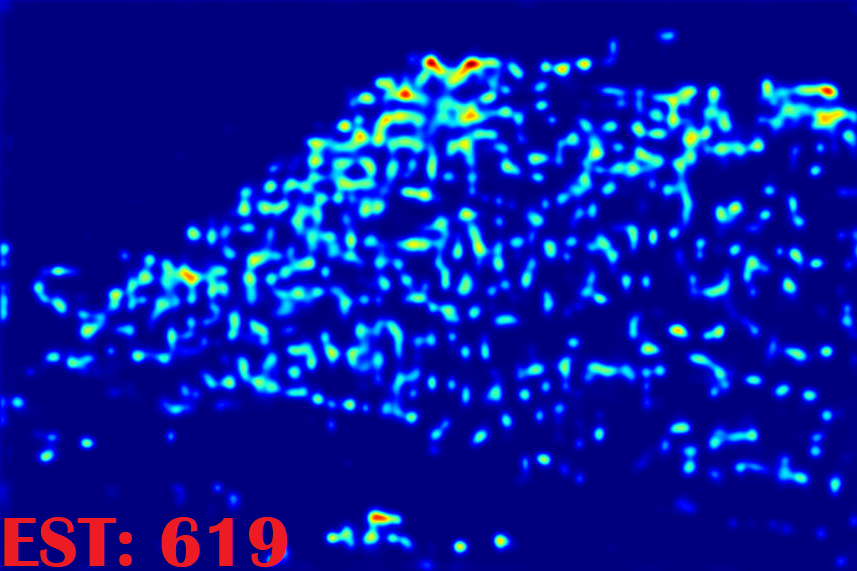}&
     \includegraphics[width=.31\linewidth]{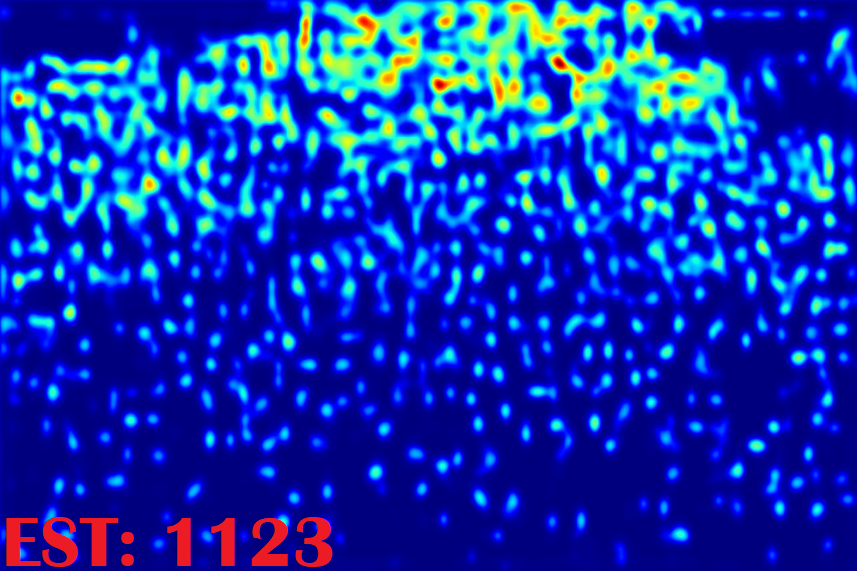}&
     \includegraphics[width=.31\linewidth]{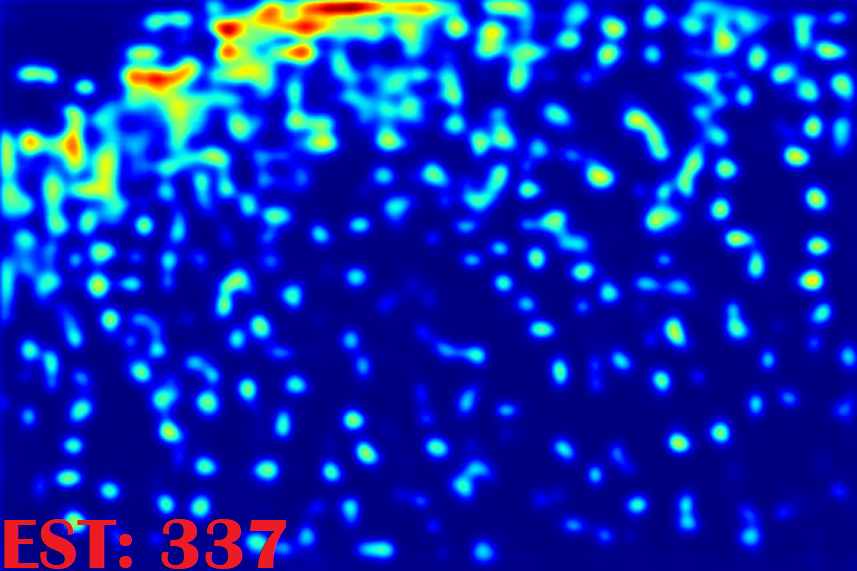}\\
     & \rotatebox[origin=l]{90}{\hspace{5mm}\footnotesize{Ours}}&
     \includegraphics[width=.31\linewidth]{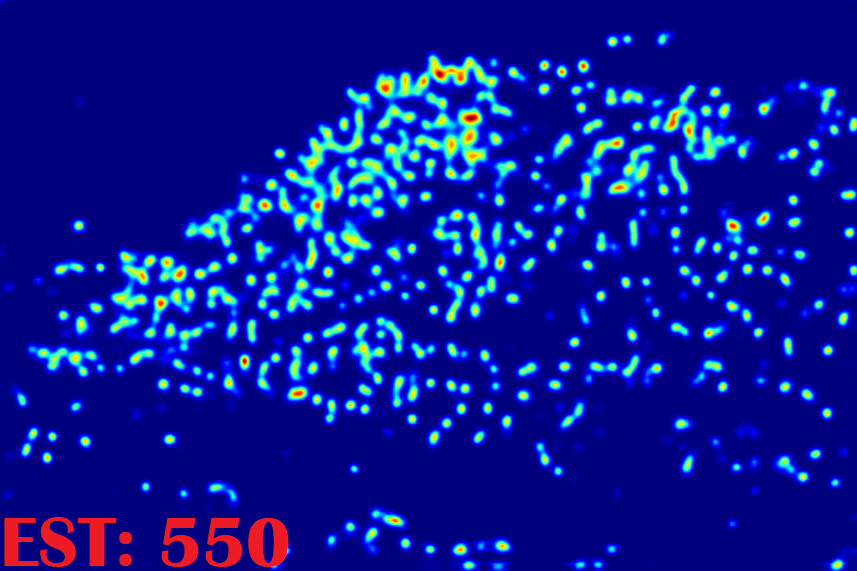}&
     \includegraphics[width=.31\linewidth]{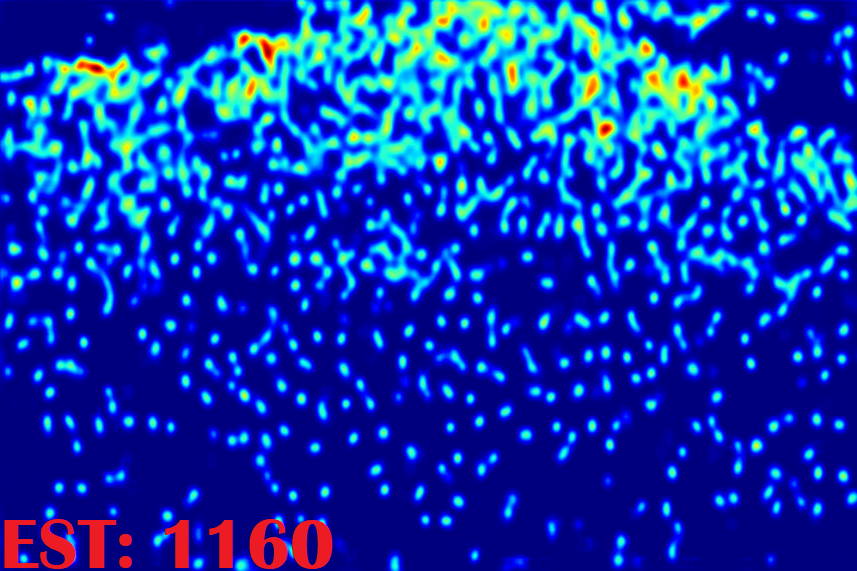}&
     \includegraphics[width=.31\linewidth]{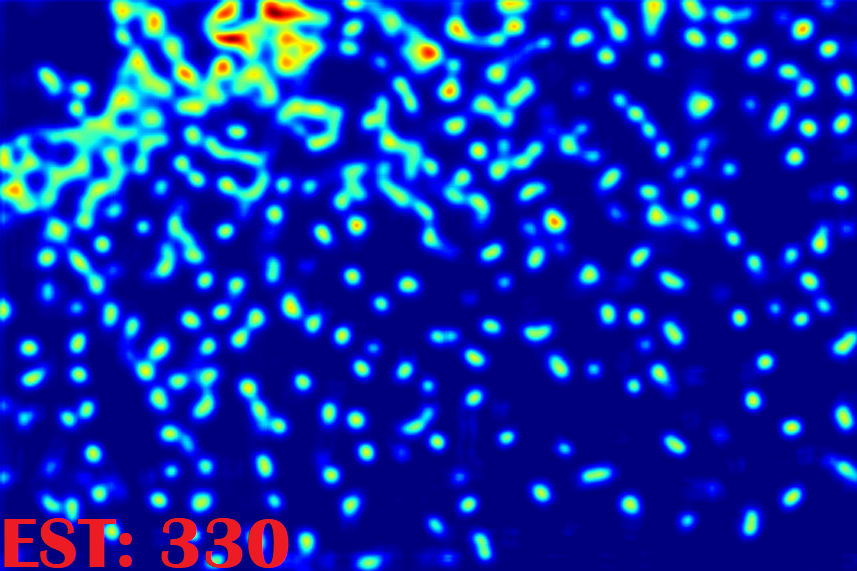}\\
     & \rotatebox[origin=l]{90}{\hspace{6mm}\footnotesize{GT}}&
     \includegraphics[width=.31\linewidth]{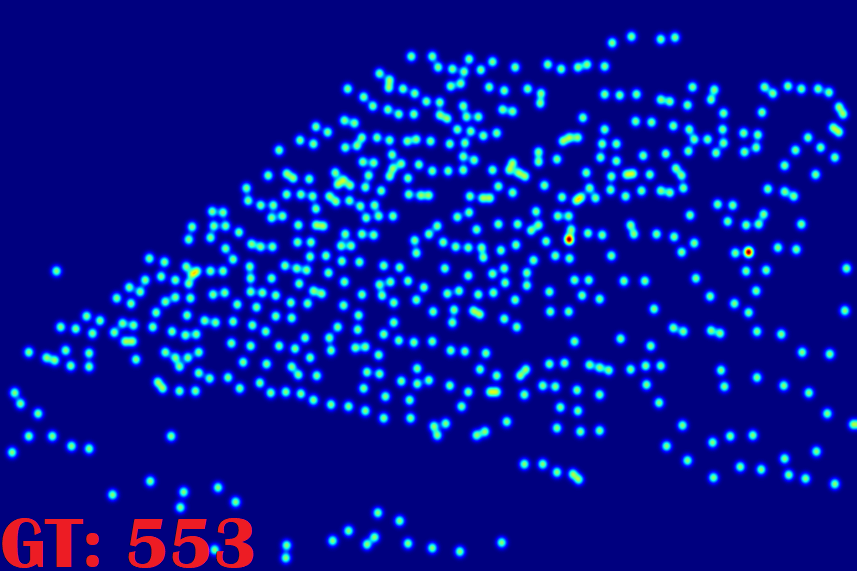}&
     \includegraphics[width=.31\linewidth]{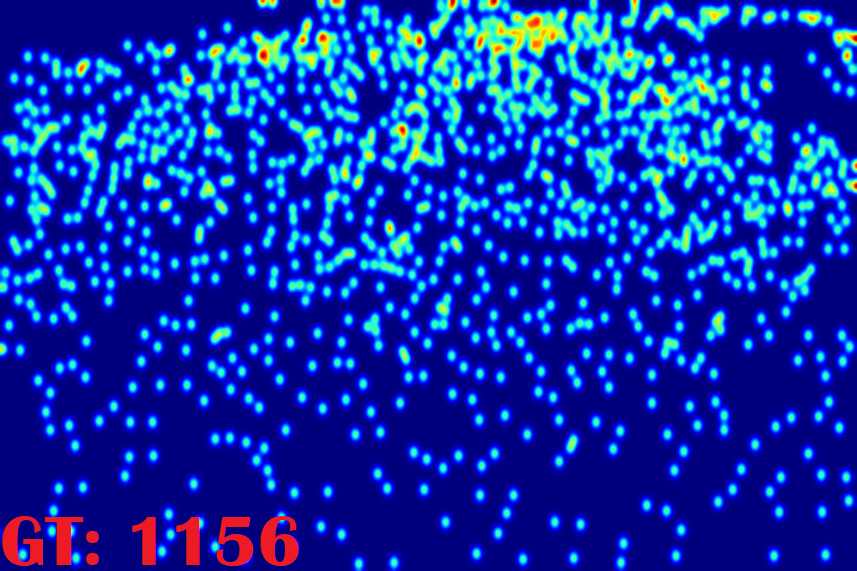}&
     \includegraphics[width=.31\linewidth]{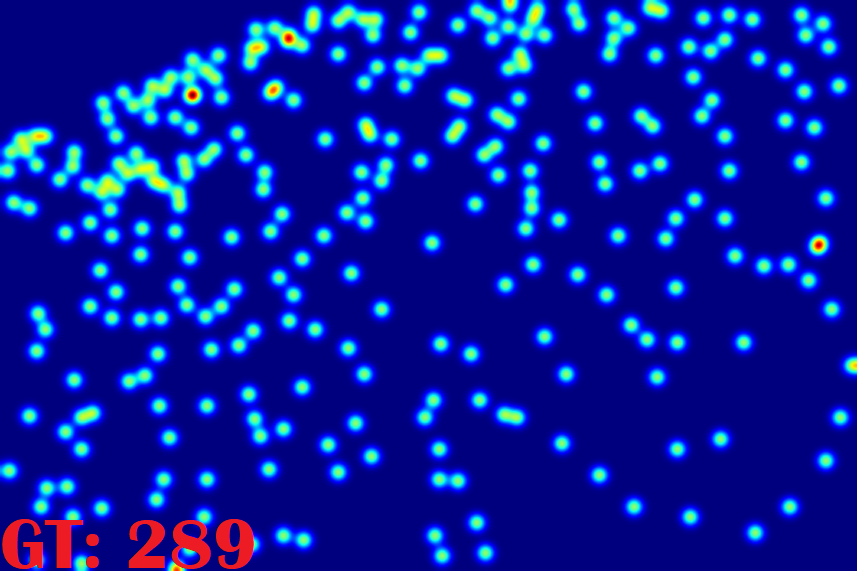}\\
  \end{tabular}
 \caption{Qualitative results of the estimated density maps in the ShanghaiTech PartA dataset with a 10\% labeling budget.}
 \label{QR figure}
 \end{figure}

Furthermore, we visualize the learned crowd affinities between the labeled and unlabeled regions in Fig~\ref{CAP figure}.
Note that labeled regions are enclosed between two red lines in each image with two representative positions marked.
Crowd affinities between each marked position and all the positions of the unlabeled regions are illustrated in ``Affinity Map $\#$''.
We can see that the affinity maps can activate the areas with the same semantics as the marked position, e.g, trees, skies, and humans with the same scale and illumination.
This indicates that supervision signals can be effectively applied to the unlabeled regions via the explicit semantic exploration brought by the CAP module.

\subsection{Comparison to State-of-the-art Methods}
In this section, we compare our method with state-of-the-art approaches, including MT \cite{tarvainen2017mean}, UDA \cite{xie2020unsupervised}, L2R \cite{liu2019exploiting}, IRAST \cite{liu2020semi}, AL-AC \cite{zhao2020active} and GP \cite{sindagi2020learning}.
Among them, MT \cite{tarvainen2017mean} and UDA \cite{xie2020unsupervised} are the widely-used generic semi-supervised methods.
L2R \cite{liu2019exploiting} is a self-supervised learning method which exploits unlabeled samples by ranking cropped pathes based on their inclusion relationships.
IRAST \cite{liu2020semi}, AL-AC \cite{zhao2020active} and GP \cite{sindagi2020learning} are the semi-supervised crowd counting methods proposed recently, which are based on the None-or-All labeling strategy.
All the comparison methods are based on CSRNet \cite{li2018csrnet} with a VGG16 backbone network.
The labeling budgets in the ShanghaiTec PartA (denoted as STPart A), ShanghaiTec PartB (denoted as STPart B), and UCF-QNRF datasets are 10\%, 10\%, and 20\%, respectively.
The comparison results are shown in Table~\ref{comparison with sota}.

We can see from Table~\ref{comparison with sota} that the general semi-supervised methods (i.e., MT \cite{tarvainen2017mean} and UDA \cite{xie2020unsupervised}) preform similarly to L2R \cite{liu2019exploiting}, and the recently-proposed crowd counting methods (i.e., IRAST \cite{liu2020semi} and AL-AC \cite{zhao2020active}) can achieve promising results compared to the general semi-supervised methods. However, they are still far from the fully-supervised CSRNet model.
Differently, our method can effectively narrow down the performance gap with the CSRNet model and enhance the state-of-the-art semi-supervised counting performance by a large margin.
For example, our method outperforms the best AL-AC~\cite{zhao2020active} by 9.4\%/8.1\% and 8.6\%/22.5\% for MAE/RMSE in the ShanghaiTec PartA and ShanghaiTec PartB datasets, respectively.
Note that as the experimental settings of GP \cite{sindagi2020learning} are different from the other methods, we have not compared with it in Table~\ref{comparison with sota}.
Following GP~\cite{sindagi2020learning}, we also annotate 5\% of the entire training set in the Shanghaitec PartA dataset.
The performance of our method is 89.7/125.6 for MAE/RMSE which is much better than 111/159 of GP.

\subsection{Qualitative Results}
Qualitative results of the estimated density maps can be seen in Fig.~\ref{QR figure}.
``None-or-All'' and ``Ours w/o MDC\&CAP'' denote the methods with the None-or-All labeling strategy and the proposed region-level strategies, respectively, and images or regions in the two methods are labeled randomly.
``Ours'' represents our method which employs the region-level labeling strategy with the proposed MDC and CAP modules.
We can see that due to the insufficient number of annotated crowd scenes, the ``None-or-All'' method can only make a coarse judgement and estimate noisy density maps with large counting errors.
From ``None-or-All'' to ``Ours w/o MDC\&CAP'', we can observe that ``Ours w/o MDC\&CAP'' can improve the density maps to some extent, but still suffers from the coarse estimations due to the limited labeling budget.
Differently, our method can achieve superior counting performance both on the estimated density maps and counting errors with the same labeling budget thanks to the proposed representative regions selection strategy and the crowd affinity propagation module.

\section{Conclusion}
In this work, we propose to break the labeling chain of previous methods and make the first attempt to reduce spatial labeling redundancy for effective semi-supervised crowd counting. Specifically, we analyze the region representativeness from both the vertical and horizontal directions, and formulate the representative regions as cluster centers of Gaussian Mixture Models based on their multi-level density vectors. Additionally, we design a Crowd Affinity Propagation (CAP) module to directly supervise the unlabeled regions via feature propagation without the error-prone pseudo label generation. Extensive experiments on widely-used benchmarks demonstrate that our method outperforms previous best approaches by a large margin.

\bibliographystyle{IEEEtran}
\bibliography{egbib}

\end{document}